\documentclass[sigconf]{acmart}
\AtBeginDocument{%
  }

\setcopyright{acmlicensed}
\copyrightyear{2018}
\acmYear{2018}
\acmDOI{XXXXXXX.XXXXXXX}
\acmConference[Conference acronym 'XX]{Make sure to enter the correct
  conference title from your rights confirmation email}{June 03--05,
  2018}{Woodstock, NY}
\acmISBN{978-1-4503-XXXX-X/2018/06}

\usepackage{amsmath,amsfonts,bm}

\def\eqref#1{equation~\ref{#1}}

\def\1{\bm{1}}

\def\ve{{\bm{e}}}

\def\vh{{\bm{h}}}

\def\vx{{\bm{x}}}

\DeclareMathAlphabet{\mathsfit}{\encodingdefault}{\sfdefault}{m}{sl}
\SetMathAlphabet{\mathsfit}{bold}{\encodingdefault}{\sfdefault}{bx}{n}

\def\gE{{\mathcal{E}}}

\def\gG{{\mathcal{G}}}

\def\gV{{\mathcal{V}}}

\newcommand{\R}{\mathbb{R}}

\usepackage{graphicx}
\usepackage{hyperref}
\usepackage{url}
\usepackage{graphicx}
\usepackage{verbatim} 
\usepackage{subcaption} 
\usepackage{amsthm}
\usepackage{svg}
\usepackage{makecell}

\newtheorem{definition}{Definition}
\newtheorem*{problem}{Problem}

\newcommand{\myparagraph}[1]{\vspace{\smallskipamount}\noindent\textbf{#1.\xspace}}

\newtoggle{showmarks}
\toggletrue{showmarks}  

\iftoggle{showmarks}{
  
  \newcommand\Fiza[1]{\textcolor{blue}{Fiza: #1}}
  \newcommand\Anjaly[1]{\textcolor{cyan}{Anjaly: #1}}
    \newcommand\Chetan[1]{\textcolor{green}{Chetan: #1}}
  \newcommand\Rujia[1]{\textcolor{yellow}{Rujia: #1}}
    \newcommand\Anson[1]{\textcolor{orange}{Anson: #1}}

  \newcommand\TODO[1]{\textcolor{red}{TODO: #1}}
}{
  \newcommand\Ayush[1]{\unskip}
  \newcommand\Fiza[1]{\unskip}
  \newcommand\Anjaly[1]{\unskip}
    \newcommand\Chetan[1]{\unskip}
  \newcommand\Rujia[1]{\unskip}
  \newcommand\Anson[1]{\unskip}
  \newcommand\TODO[1]{\unskip}

}

\newtheorem{observation}{Observation}
\usepackage{enumitem}  
\usepackage{comment}
\usepackage{algpseudocode}
\usepackage{algorithm}

\usepackage{graphicx}
\usepackage{adjustbox}
\usepackage{multicol}

\usepackage{xspace}
\newcommand{\company}{CompanyX\xspace}

\begin{document}

\title{Attention Enhanced Entity Recommendation for Intelligent Monitoring in Cloud Systems}

\author{Fiza Husain}
\affiliation{%
  \institution{}
  \country{}
  }
\email{fizahusain1110@gmail.com}
\author{Anson Bastos}
\affiliation{%
  \institution{Microsoft}
  \country{}
  }
\email{ansonbastos@microsoft.com}
\author{Anjaly Parayil}
\affiliation{%
  \institution{Microsoft}
  \country{}
  }
\email{aparayil@microsoft.com}
\author{Ayush Choure}
\affiliation{%
  \institution{Microsoft}
  \country{}
  }
\email{aychoure@microsoft.com}
\author{Chetan Bansal}
\affiliation{%
  \institution{Microsoft}
  \country{}
  }
\email{chetanb@microsoft.com}
\author{Rujia Wang}
\affiliation{%
  \institution{Microsoft}
  \country{}
  }
\email{rujiawang@microsoft.com}
\author{Saravan Rajmohan}
\affiliation{%
  \institution{Microsoft}
  \country{}
  }
\email{saravar@microsoft.com}

\begin{abstract}

In this paper, we present DiRecGNN, an attention-enhanced entity recommendation framework for monitoring cloud services at \company. We provide insights on the usefulness of this feature as perceived by the cloud service owners and lessons learned from deployment. Specifically, 
we introduce 
the problem of \emph{recommending the optimal subset of attributes (dimensions)} that should be tracked by an automated watchdog (monitor) for cloud services. 
To begin, we construct the monitor heterogeneous graph at production-scale. 
The interaction dynamics of these entities are often characterized by limited structural and engagement information, resulting in inferior performance of state-of-the-art approaches. Moreover, traditional methods fail to capture the dependencies between entities spanning a long range due to their homophilic nature. Therefore, we propose an attention-enhanced entity ranking model inspired by transformer architectures. Our model utilizes a multi-head attention mechanism to focus on heterogeneous neighbors and their attributes, and further attends to paths sampled using random walks to capture long-range dependencies. We also employ multi-faceted loss functions to optimize for relevant recommendations while respecting the inherent sparsity of the data. Empirical evaluations demonstrate significant improvements over existing methods, with our model achieving a 43.1\% increase in MRR. Furthermore, product teams who consumed these features perceive the feature as useful and rated it 4.5 out of 5. 
\end{abstract}




\maketitle
\section{Introduction}
Cloud service providers must consistently oversee their services to maintain high availability and reliability. Any lapses in monitoring can result in delayed incident detection and considerable negative impact on customers. For this purpose, automated watchdogs called \emph{monitors} are created. Monitors continuously track certain metrics, which are the time-series signals emitted by the service in specific \emph{dimensions}. These dimensions are physical or logical divisions within the service, such as environment, region, and status code. When monitors are created, they aggregate the metrics along a subset of these dimensions based on the monitor's purpose.
The existing approach to selecting these dimensions is unstructured and reactive. Developers rely on their collective experience and predominantly use a trial-and-error method. Consequently, monitors frequently suffer from incomplete coverage of dimensions, leading to misdetection of incidents, or tracking redundant dimensions, causing unnecessary noise and wasted effort. 
In this work, we pose the selection of the relevant set of dimensions needed to create a monitor as a recommendation problem.
We represent the connectivity by a graph network accounting for the heterogeneous relationships between entities such as dimensions, metrics, monitors, and their additional similarities.

Graph neural networks (GNNs) have proven to learn effective representations for a wide range of real-world systems, including social media graphs \citep{kipf2016semi,hamilton2017inductive,borisyuk2024lignn, sankar2021graph, zhang2018end,borisyuk2024lignn}. Most initial works focus on learning over homogeneous graphs where the nodes and edges are of the same type.
Heterogeneous graph neural networks (HGNN) offer the flexibility to encode both structured and unstructured information associated with various node types and to model the explicit links between different nodes and unstructured features associated with nodes, such as texts and images \cite{zhang2019heterogeneous}.
HGNNs use either message-passing to learn effective node representations from local graph neighborhoods containing structural relations among nodes and unstructured content \cite{zhang2019heterogeneous,hong2020attention,zhao2021heterogeneous,hu2019cash, hu2020heterogeneous}, or metapath-based neighbors \cite{wang2019heterogeneous, fu2020magnn, yun2019graph}.

\textbf{Graph networks for latent entity representation}: Recently, HGNNs have demonstrated promising results in several industrial systems designed for item recommendations in bipartite \cite{ying2018graph} or multipartite \cite{yang2020multisage} user-to-item interaction graphs. GCN \cite{kipf2016semi}, GraphSAGE \cite{hamilton2017inductive}, and GAT \cite{velickovic2017graph} employ various network architecture and self attention mechanism to aggregate the feature information from neighboring nodes. Further, scalable extensions to these techniques were introduced in \cite{zeng2019graphsaint,chiang2019cluster, huang2018adaptive}.   
GRAFRank \cite{sankar2021graph} extends GNNs  for large-scale user-user social modeling applications and employs multi-modal neighbor
aggregators and cross-modality attentions to learn user
representations.
Yet, entity ranking using heterogeneous graph networks with the structural and engagement properties is still a core challenge which reduces the quality of the recommended entities.
In this work, we investigate the problem of dimension recommendation for monitoring cloud services using HGNNs and propose an attention enhanced entity recommendation framework. 

\begin{figure}[t]
    \centering   \begin{subfigure}[b]{0.5\textwidth}
     \centering
\includegraphics[width=\linewidth]{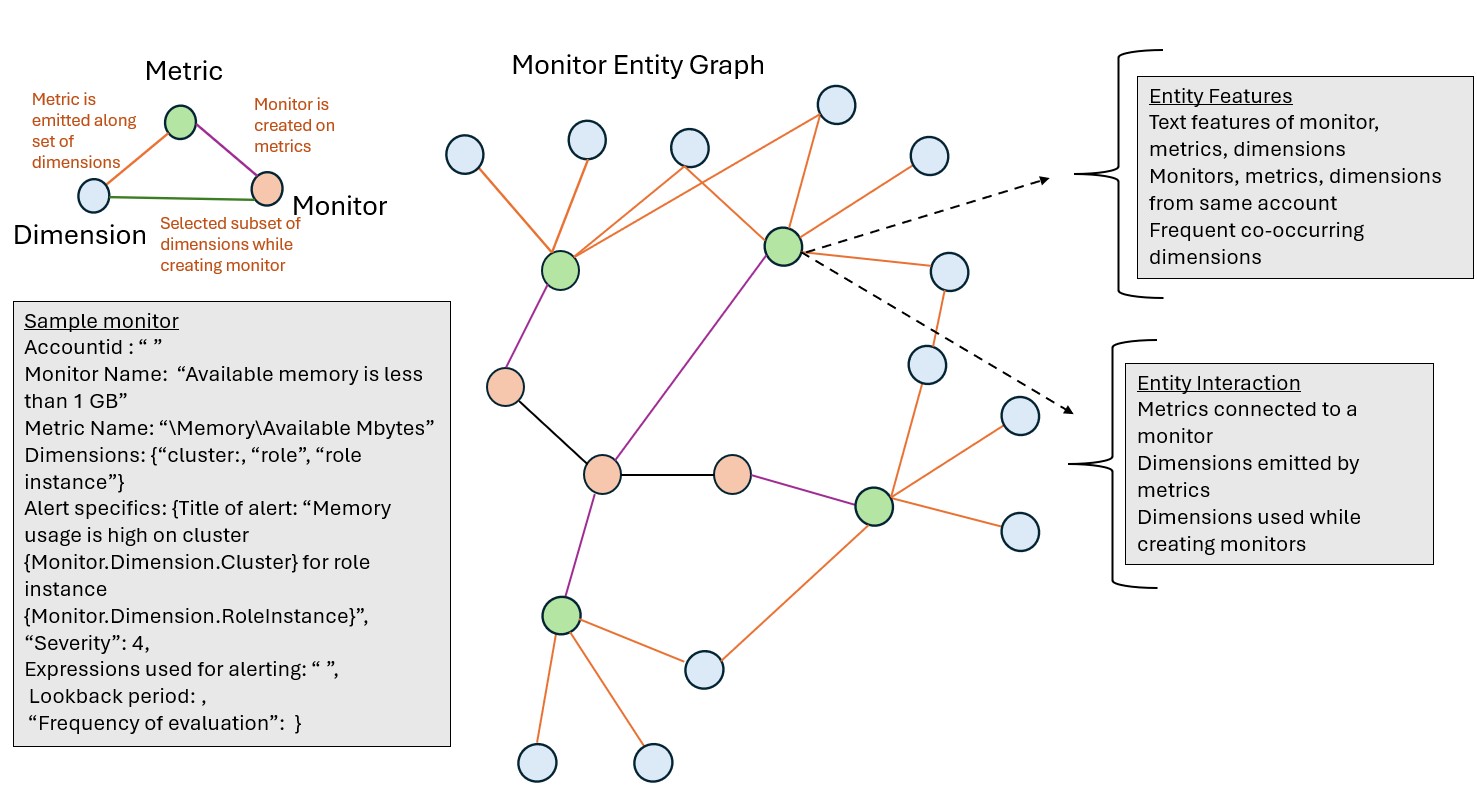}
        \caption{}
           \label{subfig:formulationa}
    \end{subfigure}%
    
    \begin{subfigure}[b]{0.5\textwidth}
         \centering
\includegraphics[width=\linewidth]{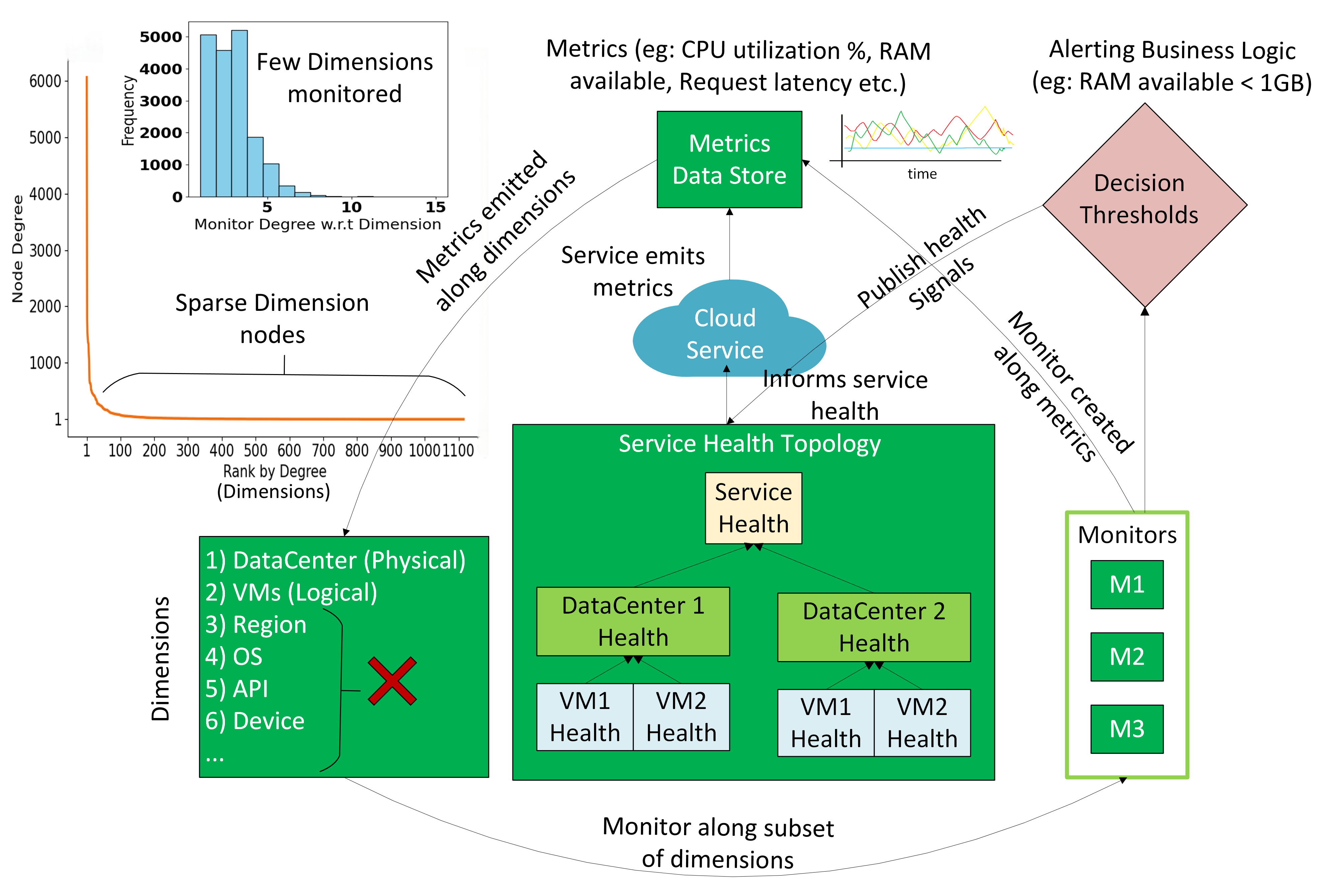}
    \caption{}
           \label{subfig:formulationb}
    \end{subfigure}
\caption{ a) Monitor entity ranking problem: Nodes represent the monitors, metrics, and dimensions in a cloud setting. Each node contains text features and interacts with their neighboring nodes using link communication and b) The overall service health where the cloud service emits metrics along many dimensions and only few are being used for monitoring health. For eg. in this case the service health depends on health of each datacenter using the service which in turn depends on the health of the individual VMs in the datacenter. 
}
\label{fig:formulation}
\end{figure}

\noindent \textbf{Entity ranking in cloud setting}:  
The \emph{monitor entity graph} consists of heterogeneous relationships between entities such as the \emph{monitors}, the \emph{metrics} (timeseries data) they emit and the \emph{dimensions} over which the metrics are monitored (cf. \autoref{subfig:formulationa}).
Recommending attributes for aggregating time-series signals to create automated watchdogs (monitors) that ensure continuous service availability is a complex problem in the cloud setting \cite{surianarayanan2019essentials,chen2020towards, montes2013gmone}. Previous research has focused on recommending time-series signals to be associated with automated watchdogs \cite{nair2015learning,srinivas2024intelligent}. Generating recommendations in a cloud setting requires leveraging domain knowledge and interaction between multiple entities. Therefore, HGNNs are a natural choice due to their effective message passing mechanisms and ability in capturing intricate relationships between entities with different types of relationship. 
However, applying HGNNs directly to cloud domain datasets is challenging. This is due to the limited structural information available from interactions which could span over long ranges and the sparse nature of the graph. Most attributes are not connected to all types of nodes, exhibit fewer degrees, and have fewer interactions. 
 As shown in \autoref{subfig:formulationb} (top-left),
 a monitor entity graph exhibits a characteristic long tail, with most monitors connected to few dimensions. Thus, in this work we focus on modeling dimensions in a \emph{monitor entity graph}, to learn robust node embeddings with limited
 structural connectivity. 
\\
\textbf{Present work}: In this work, we target the problem of finding the optimal subset of \emph{dimensions} (out of the universal set of all available dimensions) that should be tracked by a \emph{monitor}. 
We pose the problem as ranking dimensions for a monitor in a recommendation framework over the \emph{monitor entity graph}. 
We begin by discussing the cloud setting and introducing the different entities involved in the recommendation problem. We then introduce the monitor entity graph and formulate the problem. Next, we discuss the specific nature of the graph, including its long-tailed distribution and the sparsity of interactions. To overcome the challenges of structural and interaction sparsity, we propose leveraging the available set of node attributes and multi-type interactions, and present an attention-enhanced entity ranking framework. 
Specifically, we make the following contributions.
\\ 
\textbf{Contributions}: 
(1) We introduce the problem of selecting relevant \emph{dimensions} for cloud \emph{monitors} and pose it as a recommendation task over a heterogeneous \emph{monitor entity graph}. Subsequently, we propose, a representation learning framework for \underline{Di}mension \underline{Rec}ommendation (dubbed DiRecGNN), enhanced with transformer-style graph convolutions.
(2) Furthermore, to overcome the limitation of the message passing framework to capture global interactions in the monitor entity graph, the framework incorporates dependencies spread over long ranges using paths obtained by sampling random walks with restart from the target nodes.
(3) Moreover, from the optimization perspective, we propose an attention-alignment loss function, which aggregates and attends to relevant sparse information from different nodes and their interaction patterns. 
(4) Finally, we run extensive experiments on the monitor-entity dataset and show significant improvements in the hit-rate, mean reciprocal rank, and recall over baselines. 
(5) 
The real world case studies show that our proposed method provides relevant recommendations (4.5/5) of dimensions for the monitors used in production. We also include lessons learned from surfacing these recommendations to service owners, which involved challenges in developing engineering solutions for retrieving and updating recommendations at scale due to their volume. Additionally, we highlight other user feedback, such as the call for end-to-end automation.

\section{Preliminaries}
In this section, we introduce the problem of recommending dimensions for creating monitors for cloud services.
We formally define the problem and provide preliminary notations.
\subsection{Problem Formulation}
Ensuring continuous availability of services is essential for cloud service providers. Cloud services continuously record information about their health in the form of run-time telemetry. This telemetry serves as signals to be analyzed for detecting anomalies. Therefore, cloud services are equipped with automated watchdogs, also known as \emph{monitors}, that monitor service health. \autoref{subfig:formulationa} illustrates the concept of a monitor and connected entities such as metrics and dimensions. 
The figure shows an example of a specific monitor, including the dimensions and metrics used by the monitor. Examples of these dimensions include indicators such as the success of an operation, file path, environment of service deployment, and the identifier of a compute node. Monitors aggregate the metrics emitted along various dimensions, and the alerting conditions operate on the aggregated signal to create an alert. A monitor also encompasses additional entities needed to create an alert, such as alerting conditions, expressions upon which the alert is created, and the frequency at which the conditions should be checked. All of these depend on the metrics and dimensions used to create a monitor.

\noindent There are several approaches to predict metrics to be monitored for a given cloud service  \cite{srinivas2024intelligent}. However, given the metric, \(m_i\) and the set of dimensions along which the metric is already being emitted, the problem of ranking dimensions along which the metric needs to be aggregated to raise an alert has not yet been explored (subset of dimensions). 
Furthermore, the monitor entity graph containing monitors, metrics, and dimensions is a heavy-tailed sparse graph, with limited interaction between most of the monitors, metrics, and dimensions (c.f. \autoref{subfig:formulationb}). 
As a result, our work focuses on developing a solution to the challenging problem of \emph{recommending dimensions for monitor creation} in the context of this heavy-tailed sparse graph. \autoref{subfig:formulationa} shows the example of features associated with each node.
We start by modeling the monitor entities and their various interactions, as well as their textual node attributes, as a heterogeneous multi-type interaction graph. From this graph, we generate effective node representations, which are then used for dimension ranking.
Next, we define the monitor entity graph and related attributes.

\begin{definition}{(Monitor Entity Graph):}
We represent the data as a heterogeneous graph \(\displaystyle \gG =(\displaystyle \gV, \displaystyle \gE) \) where \(\displaystyle \gV=\{\displaystyle \gV_m \cup \displaystyle \gV_d \cup \displaystyle \gV_k\}\) represents the set of nodes with \(\displaystyle \gV_m, \displaystyle \gV_d, \displaystyle \gV_k\) denoting monitors, dimensions, and metrics, respectively and \(\displaystyle \gE=\{\displaystyle \gE_{md} \cup \displaystyle \gE_{kd} \cup \displaystyle \gE_{mk}\}\) represents the set of edges, capturing three types of relationships:
1) \(\displaystyle \gE_{md}\): ``monitor associated with dimension'', 2) \(\displaystyle \gE_{kd}\): ``metric has dimension'', and 3) \(\displaystyle \gE_{mk}\): ``monitor emits metric''.
\end{definition}
The graph structure \(\gG\) captures the intricate relationships in the data that are crucial for making informed recommendations. By explicitly modeling different entity types and their relationships, we enable the model to learn domain-specific patterns. Each node in the graph, denoted as \(v \in \gV\), has a unique initial representation given by \(\vx_v \in \mathbb{R}^d\), where \(d\) is the dimension of the embedding space. The vector \(\vx_v\) is the concatenation of two types of features: intrinsic features, which are domain-specific attributes of the entity (e.g., metric name, dimension name, monitor name, related service), and learned embeddings, which are trainable embeddings that capture the entity's role in the graph structure (e.g., co-occurrence with another node). The monitor entity graph is shown in \autoref{subfig:formulationa}.  
It is to be noted that we assume a static graph.

 We define the problem of ranking dimensions for monitor creation in the monitor-entity graph,  \(\gG\) with different types of node features and link features as follows:
\begin{problem}[Dimension Ranking using Heterogeneous Interactions]
Leverage  entity features \{\(\displaystyle \vx_v: v \in    \displaystyle \gV \) \}, link features $\{ \displaystyle \ve_{v_1v_2} \in  \displaystyle \gE\}$ and monitor entity graph, $\displaystyle \gG$, to generate entity representations that facilitate relevant recommendations of dimension nodes ($v \in V_d$) for a given monitor node ($v \in V_m$).
\end{problem}

\subsection{Definition of heterogeneous graph and notations} \label{apndx:hetero_prelims}
We introduce the basic notations and formulations useful during message passing in GNN frameworks. GNNs use multiple layers to learn node representations. At each layer $l > 0$ (where $l = 0$ is the input layer), GNNs compute a representation for node $v_1$ by aggregating features from its neighborhood through a learnable aggregator $F_{\theta,l}$ per layer \cite{hamilton2017inductive, kipf2016semi, velickovic2017graph}. The embedding for node $v_1$ at the $l$-th layer is given by:

\begin{equation}
    \displaystyle \vh_{v_1,l} = F_{\theta,l}( \displaystyle \vh_{v_1,l-1}, \{ \displaystyle \vh_{v_2,l-1}\}), v_2 \in N(v_1)
\end{equation}

The embedding $\displaystyle \vh_{v_1,l}$ at the $l$-th layer is a non-linear aggregation of its embedding $\displaystyle \vh_{v_1,l-1}$ from layer $l - 1$ and the embedding of its immediate neighbors $v_1 \in N (v)$. The function $F_{\theta,l}$ defines the message-passing function at layer $l$ and can be instantiated using a variety of aggregators \cite{hamilton2017inductive, kipf2016semi, velickovic2017graph}. The node representation for $v_1$ at the input layer is $ \displaystyle \vh_{v_1,0}$, where $\displaystyle \vh_{v_1,0} = \displaystyle \vx_{v} \in  \displaystyle \R^{D}$.

In order to capture higher order relations in the heterogeneous graphs, one can use the notion of meta-paths which we define next.
\begin{definition}{(Meta-path):}
 In a heterogeneous graph the nodes can be connected by relations of different types. A meta-path is a path following a defined schema of specific relation types connecting certain node types. A meta-path $P$ on the heterogeneous graph can be denoted as $v_1 \xrightarrow{r_{12}} v2 \xrightarrow{r_{23}} v_3 \ldots \xrightarrow{r_{(m-1)m}} v_m$, where $v, r$ are the node and edge types respectively and $(m-1)$ is the path length.
\end{definition}
While meta-paths are useful to capture long range dependencies, in production scale heterogeneous graphs however, there could be an exponential number of these paths with the path length, which makes the use of all meta-paths intractable. Thus we propose to sample these meta-paths using random walks with restart from the target node. As of now we explore random sampling of these paths and future works could consider the data sparsity pattern to do a weighted sampling of paths.

\section{Data Description and Insights} \label{data_desc}

In this section, we analyse the monitor entity graph and discuss insights into the structure of the network, as well as the features that impact node relationships. We then use these insights to inform the framework design for dimension recommendations and ranking.

\begin{figure}[h!]  
\centering
\begin{subfigure}[b]{0.22\textwidth}
       \centering
\includegraphics[width=1.0\linewidth]{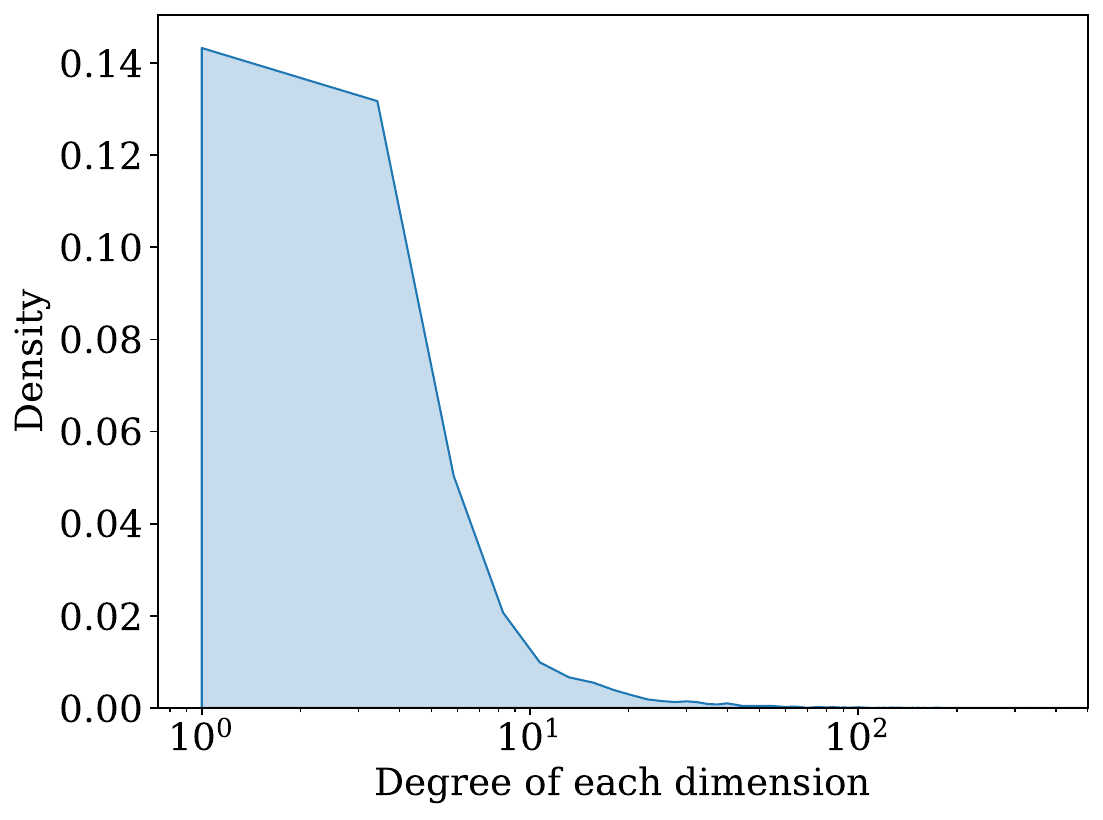}
\caption{}
\label{subfig: dimpopularity2}
\end{subfigure}%
\begin{subfigure}[b]{0.22\textwidth}
       \centering
\includegraphics[width=1.0\linewidth]{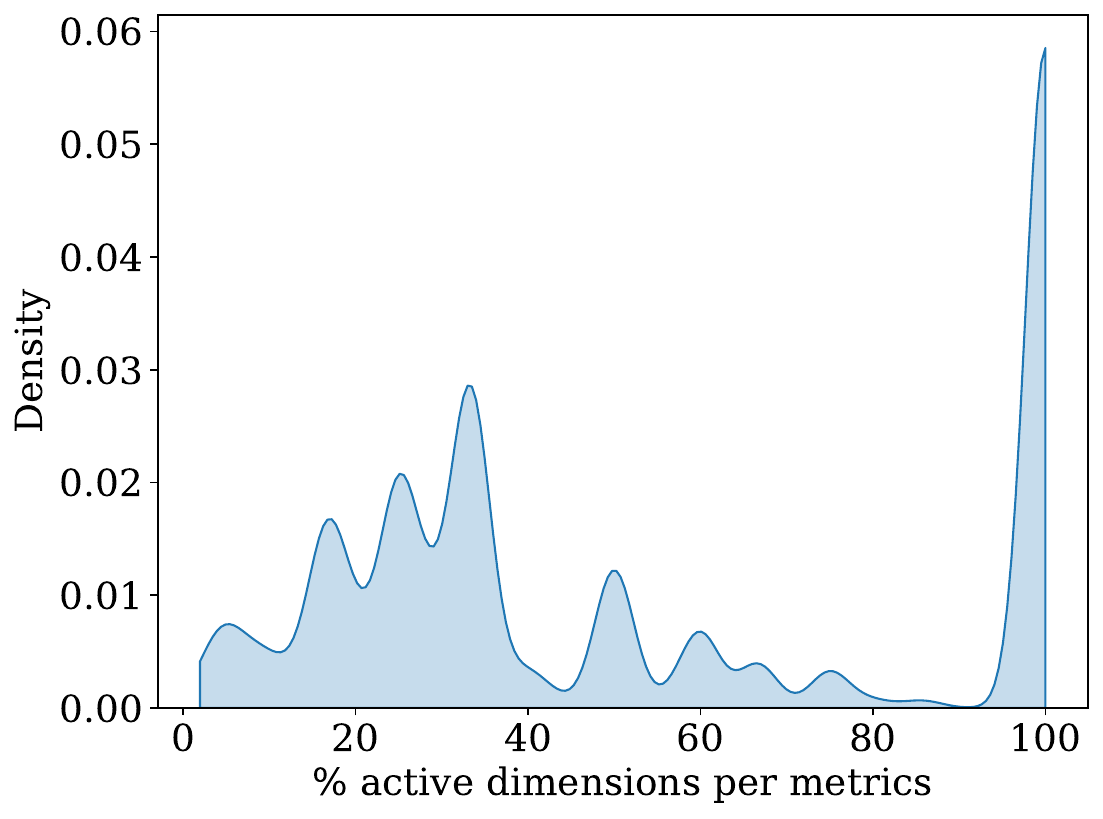}
\caption{}
\label{subfig: dimpopularity3}
\end{subfigure}
    \caption{
    Characteristics of the Monitor Entity Graph: a) Distribution of dimension degree based on the metric-to-dimension interactions from the monitor entity graph, and b) Distribution of the percentage of dimensions selected from the set of all dimensions along which the metric is emitted. 
    }
    \label{fig: dimpopularity}
\end{figure}

\autoref{fig: dimpopularity} illustrates the characteristics of the monitor entity graph.  
\autoref{subfig: dimpopularity2} shows the distribution of dimension degree based on the metric-to-dimension interactions from the monitor entity graph. The likelihood test ratio indicates a resemblance to long-tailed distributions. 
Next, we analyze the distribution of the percentage of dimensions selected, for monitoring, from the set of all dimensions along which the metric is emitted (observed). As seen in \autoref{subfig: dimpopularity3}, the distribution is skewed to the left, indicating that the majority of monitors (94\%) do not need to aggregate the metrics along all dimensions along which they are emitted. 
Thus the problem of selecting an optimal subset of dimensions on which to monitor is relevant.

\begin{observation}
The ``monitor entity'' graph exhibits 
activity sparsity. Although many dimensions are associated with metrics, only a subset of them is used to aggregate the metric while raising an alert.
\end{observation}
Next, we analyze the features associated with the nodes to understand its impact on recommendations. We start with the text features associated with the monitor entity graph (c.f. \autoref{subfig:formulationa}) and its effect on dimensions associated with the monitors.
\begin{figure}[h!]  
\centering
\begin{subfigure}[b]{0.22\textwidth}
       \centering
\includegraphics[width=\linewidth]{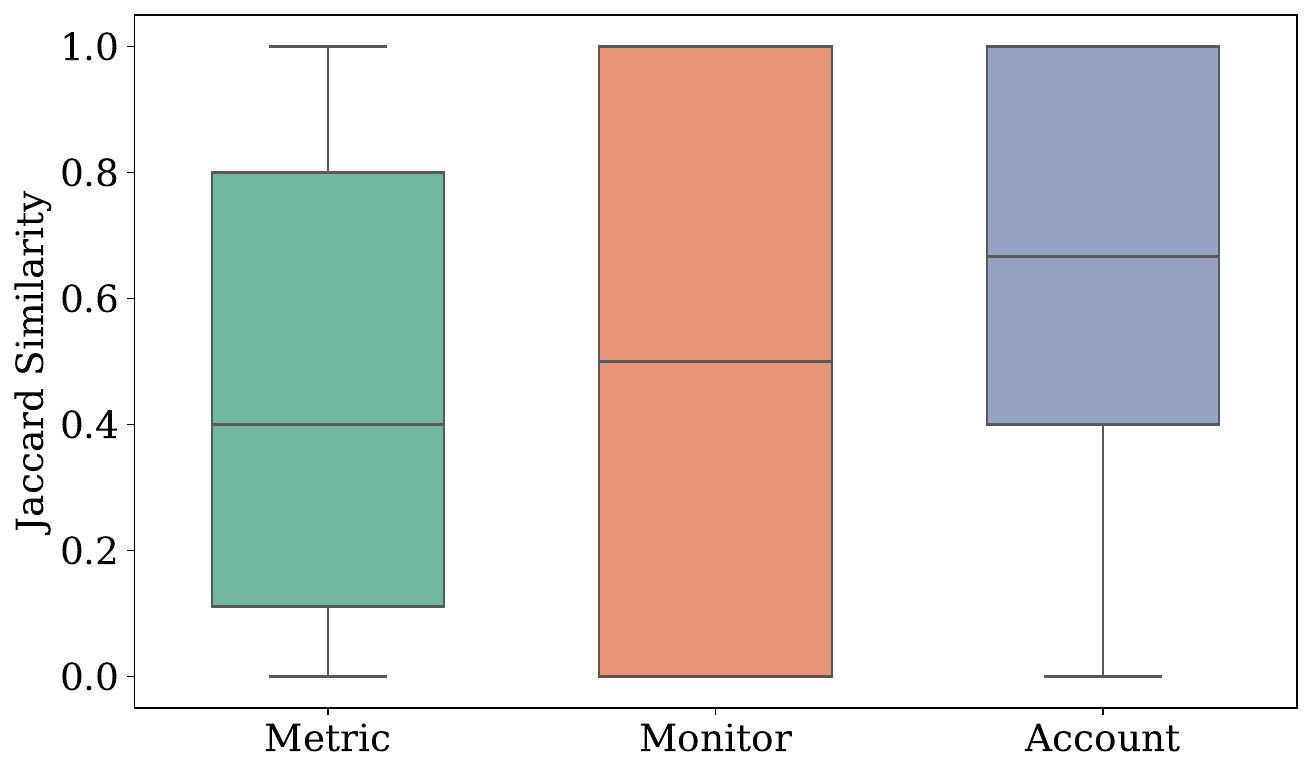}%
\caption{}
\label{subfig: graphanalysis1}
\end{subfigure}
\begin{subfigure}[b]{0.22\textwidth}
       \centering
\includegraphics[width=\linewidth]{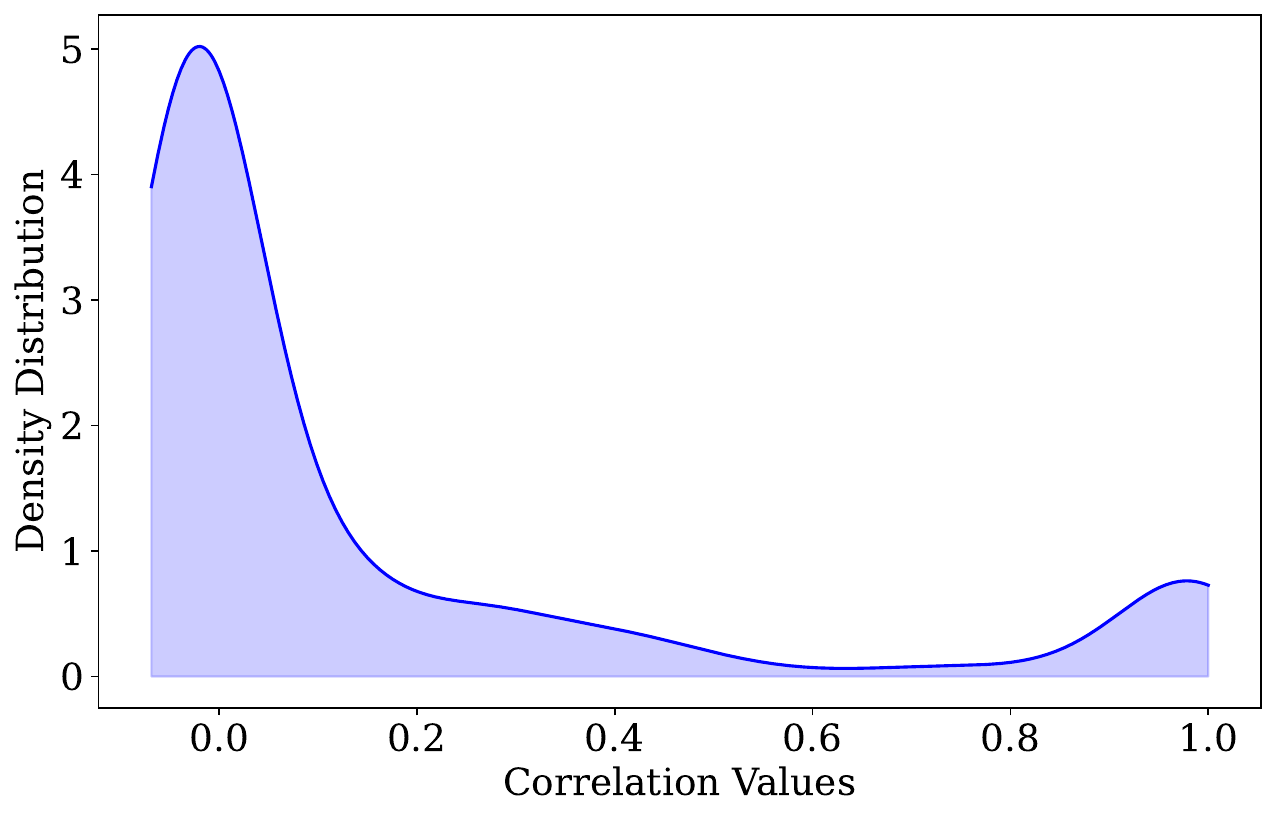}
\caption{}
\label{subfig: graphanalysis2}
\end{subfigure}
    \caption{a) Variation in jaccard similarity of set of dimensions associated with monitors with similar metric, monitor names, and same service account, and b) Distribution of pairwise correlation between dimensions. 
    }
    \label{fig: graphanalysis}
\end{figure}
\autoref{subfig: graphanalysis1}
shows the distribution of Jaccard similarity between sets of dimensions associated with monitors that exhibit high cosine similarity (\(> 0.8\)) between different text features. We consider the text similarity in the monitor names, metric names, and the service account associated with the monitor. The feature embeddings are generated using the "E5" embedding model, a general-purpose model trained through contrastive learning \citep{wang2022text}. The Jaccard similarity of dimensions from monitors exhibits different trends with respect to the similarity in the metric, monitor names, and that from the same service account. Furthermore, the similarity in dimensions with similar monitor names shows higher variance. In addition, \autoref{subfig: graphanalysis2} shows the density of pairwise correlation between dimensions connected to a monitor. The correlation plot shows two peaks signifying the presence of distinct groups with different trends. As seen in the figure, the second peak signifies the co-occurrence of a specific set of dimensions.
\begin{observation}
We observe a significant overlap in the dimensions used by monitors associated with the same service account, those with similar metrics, and those with similar names. However, the extent of similarity varies across the features. Further, the correlation between dimension pairs shows the presence of two distinct groups, where some dimensions are positively correlated, while the other group is not correlated.
Specifically, the framework  for representing entities of a node should consider the varying degrees of similarity across different features and the distinct correlation patterns among dimension pairs.
\end{observation}

\section{Attention Enhanced Entity Ranking for Sparse Cloud Graphs}\label{method}

\begin{figure*}[h!]
    \centering
    \includegraphics[width=0.95\textwidth]{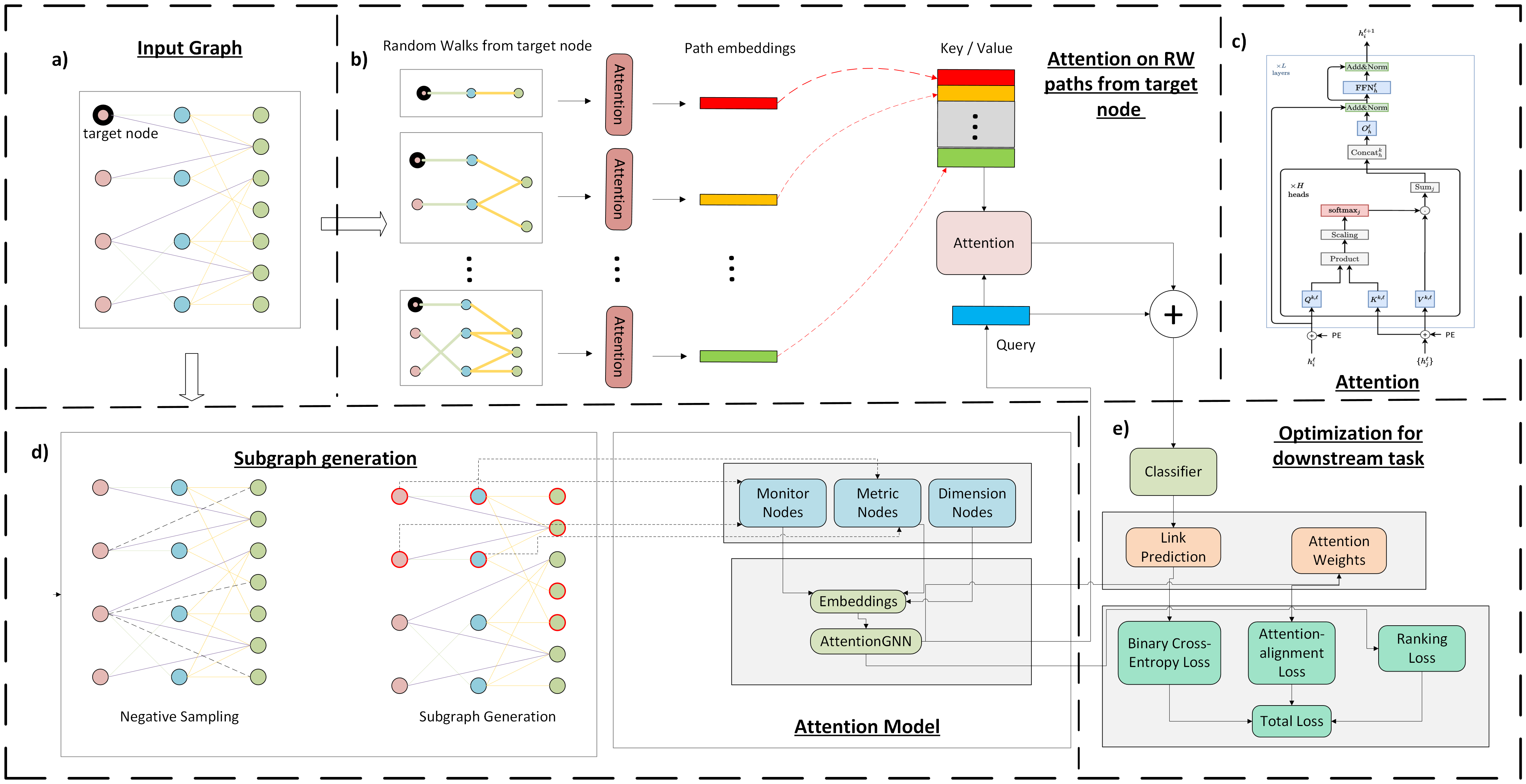}
    \caption{ The overall architecture of DiRecGNN framework. The framework uses an enhanced transformer-style graph convolution over the input graph (block a) which is sampled (block d). Our method incorporates multi-head attention mechanisms (block c), enabling the model to capture the graph structure. Note position encodings (PE) are added for sequence attention but not in case of sparse graph attention. Additionally, we sample random walks (block b) from the target node in order to better capture complex long-range dependencies. The custom losses (block e) include attention-alignment, ranking loss and BCE loss.}
    \label{fig:framework}
\end{figure*}

In this section, we present the framework to encode the multifaceted nature of cloud monitoring data and generate node representation for different entities. The graph representation, \(\displaystyle \gG \)  captures both the inherent properties of each entity and its context within the graph structure. The domain-specific attributes provide an inductive bias based on domain knowledge, while the learned embeddings allow the model to discover latent relationships and characteristics generalizing to new entities not seen during training.
The incorporation of paths from random walks allow us to capture the long range dependencies inherent in the dynamic cloud environments that may be lost during message passing \cite{balcilar2021analyzing}.
\autoref{fig:framework} shows the overview of our framework.
First we discuss the messaging passing mechanism.

\subsection{Message Passing Mechanism}
The framework employs an effective message passing mechanism designed to capture complex relationships and contextual information (\autoref{fig:framework} block c,d).
The multi-head attention enhanced message passing approach leverages edge-specific transformations and heterogeneous neighborhood aggregation. The key components of our message passing mechanism are:

\myparagraph{Multi-Head Attention} \label{sunsec_mha}
A multi-head attention mechanism to allow the model to focus on different aspects of node relationships simultaneously. This is important for distinguishing the relevance of different neighbors and relationship types in our graph. The attention weights \(\alpha\) are computed as follows:
$\alpha_{ij}=\frac{(q_i \cdot k_j)}{\sqrt{d_h d_o}}$,
    where \(q_i\) is the query vector for node \(i\), \(k_j\) is the key vector for node \(j\), \(d_h\) is the number of attention heads, and \(d_o\) is the output dimension per head.
    The attention weights are then normalized using softmax:
    $\alpha_{ij} = \frac{\exp(\alpha_{ij})}{\sum_{k \in \mathcal{N}(i)} \exp(\alpha_{ik})}$ 
    where \(\mathcal{N}(i)\) is the set of neighbors of node \(i\).
      
 \myparagraph{Edge-Aware Message Transformation} The message passing considers the type of relationship between the nodes. The transformed message incorporates both the node features and the computed attention weights, ensuring that the message is sensitive to the specific relationship type. The transformed message \(m_{ij}\) from node \(j\) to node \(i\) is computed as $m_{ij} = \alpha_{ij} v_j $, where \(v_j\) is the value vector of node \(j\). The final aggregated message for node \(i\) is $m_i = \sum_{j \in \mathcal{N}(i)} m_{ij}$. The node features are updated as: \newline 
 \begin{equation}\label{eq_mha}
      x_i^{(l+1)} = \sigma\left(W^{(l)} \cdot \text{CONCAT}\left(x_i^{(l)}, m_i^{(l)}\right)\right)
 \end{equation}
 where \(x_i^{(l)}\) is the feature vector of node \(i\) at layer \(l\), \(W^{(l)}\) is the learnable weight matrix for layer \(l\), \(\sigma\) is the ReLU activation function and CONCAT is the concatenation operation.

By using attention mechanisms, our approach can dynamically assign importance to different types of relationships and neighbors, important for our problem. The multi-head attention allows the model to leverage limited interactions more effectively, mitigating the impact of sparse data.

\subsection{Incorporating Random Walk Paths to Capture Global Properties}
While the attention mechanism described in the above section is able to capture the graph structure, it has been shown in previous works \cite{balcilar2021analyzing, bastos2022how}, that the attention mechanism is only able to effectively learn homophilic signals. This means that these models are only able to learn the neighborhood information while ignoring the long range global dependencies. In order to capture global dependencies effectively we explicitly incorporate the long range information in the form of random walk paths from the target nodes (\autoref{fig:framework} block b).
Specifically, for a given monitor node $m_i$ we have the meta paths $\{p^1 = \left( m_{i1}^1 \xrightarrow{} k_{i1}^1 \xrightarrow{} d_{i1}^1 \xrightarrow{} k_{i2}^1 \ldots d_{il}^1 \right), p^2 = \left( m_{i1}^2 \xrightarrow{} k_{i1}^2 \ldots d_{il}^2 \right) \ldots p^M = \left( m_{i1}^M \xrightarrow{} k_{i1}^M \ldots d_{il}^M \right)  \}$, where $L=2+4(l-1)$ is the length of the path and $M$ are the number of paths of length L between the target node ($d_i$) and the destination node type (dimension in this case). Due to the exponential number of possible paths with increase in path length, we sample a fixed number of paths $W$ by performing random walks of length $L$ starting from the target node. Formally, the random walk operator (for 1 hop) can be expressed as $RW = D^{-1}A$ where $D$ is the diagonal degree matrix and $A$ is the graph adjacency matrix. For k-hops, it can be computed by $RW^{k}$ and for path length $L$ for node $v$ we can sample nodes along the path from $\{RW^{i}[v]\}_{i=\llbracket1..N\rrbracket}$. This enables us to scale to large monitor entity graphs while capturing the long term dependencies. In order to capture the node signals along the path without loss of information as in message passing frameworks, we apply the attention mechanism over the sequence of nodes in the path for each path. Formally for path $p^i$, the information over the path is captured in the embedding $e_{p^i}$ as follows: 
$e_{p^i} = softmax \left(\frac{Q^i \cdot K^i}{\sqrt{d_h d_o}} \right)V^i $
    where \(Q^i (=W^Q E^i), K^i (=W^K E^i), V^i (=W^V E^i) \in R^{(L+1) \times d}\) are the query, key and value matrices for nodes along path \(i\), \(W^Q\), \(W^K\), \(W^V\) are the respective linear transformation matrices, \(d_h\) is the number of attention heads, and \(d_o\) is the output dimension per head, $E^i \in R^{(L+1) \times d}$ represents the node embeddings added with position embeddings \cite{vaswani2023attentionneed} for the sequence in the path. 
Once we have the path embeddings $P_{x_i} = \{ e_{p^i} | \quad i\in \{0,1\ldots W-1\}\}$ for a given target node $x_i$, we proceed to compute the attention with the query embedding being the output of the attention mechanism $x_i^{(l+1)}$ defined in equation \ref{eq_mha} with $P_{d_i}$ as the key, value embeddings. The final aggregated embedding (Random Walk Attention or RWA) from the random walk paths for node $x_i$ is given by:
\begin{equation}\label{eq_rwa}
     RWA_{x_i} = softmax \left(\frac{x_i^{(l+1)} \cdot P_{x_i}^T}{\sqrt{d_o}} \right) P_{x_i}^T
\end{equation}

The outputs from equations \ref{eq_mha} and \ref{eq_rwa} are then concatenated and given to the further layers for the downstream tasks.
After multiple rounds of message passing and after incorporating monitor-dimension paths, each monitor node has an enriched representation incorporating information from relevant dimensions and metrics. These representations are used to score potential monitor-dimension pairs. 
\subsection{Training Objective}
In recommendations systems, the choice of training objective directly influences the model's ability to make accurate and well-ranked recommendations. We define a composite loss function that addresses multiple aspects of the recommendation task simultaneously (\autoref{fig:framework} block e). 
Our loss function combines three key components: 1) Binary Cross-Entropy Loss, 2) TOP1-max Ranking Loss, and 3) Attention-alignment Loss.

\textbf{Binary Cross-Entropy Loss}: The binary cross-entropy (BCE) loss is fundamental for our basic prediction accuracy which is defined as:
\begin{equation}
    \mathcal{L}_{\text{BCE}} = -\frac{1}{N} \sum_{i=1}^N \left[ y_i \log(\hat{y}_i) + (1 - y_i) \log(1 - \hat{y}_i) \right]
\end{equation}
where \(y_i\) is the true label and \(\hat{y}_i\) is the predicted probability. 
The BCE loss ensures basic prediction accuracy.

\textbf{TOP1-max Ranking Loss}: To optimize the order of recommendations, we use the TOP1-max ranking loss proposed in \cite{hidasi2018recurrent}. This loss aims to push the target score above the scores of negative samples while also acting as a regularizer. It's defined as:
\begin{equation}
    \mathcal{L}_{\text{TOP1-max}} = \sum_{j=1}^N  s_j \left[ \sigma(r_j - r_i) + \sigma(r_j^2) \right]
\end{equation}
where $r_i$ is the score for the positive sample, $r_j$ are scores for negative samples, $s_j = \text{softmax}(r_j)$, and $\sigma$ is the sigmoid function.
The first term of the TOP1-max ranking loss encourages the positive score to be higher than the maximum negative score and the second term acts as a regularizer, pushing negative scores towards zero. This loss optimizes the order of recommendations, crucial for top-k recommendation scenarios.

\textbf{Attention-alignment Loss:}  
In our problem of recommending dimensions we would want the model to focus on few entities in order to recommend the relevant dimensions which, as we have seen in section \ref{data_desc}, are generally sparse.
One empirical finding is that multiple heads focusing on different entities deteriorates performance. Hence, to encourage the model to focus on few entities and to keep the attention sparse, respecting the graph structure, we introduce an attention-alignment loss component. 
The attention-alignment loss (\(\mathcal{L}_{\text{al}}\)) aims to align the attention values across different heads and is defined as:
     \begin{equation}
    \mathcal{L}_{\text{al}} = \lambda_{\text{al}} \cdot \frac{1}{L} \sum_{l=1}^L \text{MSE}(\alpha_l, \bar{\alpha}_l) \label{eqn:divloss}
 \end{equation}
  
     where $\alpha_l$ are the attention weights in layer $l$, $\bar{\alpha}_l$ is their mean, and $\lambda_{\text{al}}$ is the alignment strength. 
     This attention-alignment loss penalizes attention patterns that diverge too much from the mean across different heads, promoting a sparse representation of the graph structure, as desired.
 It prevents the model from focusing on too many entities, particularly important in sparse cloud data settings.

Our final loss function combines all the above components. 
\begin{equation}
    \mathcal{L}_{\text{total}} = \mathcal{L}_{\text{BCE}} + \mathcal{L}_{\text{al}} + \mathcal{L}_{\text{TOP1}}
\end{equation}

\textbf{Dynamic Loss Balancing}
To ensure optimal contribution from each loss component, we use a dynamic loss balancing mechanism which adjusts the weights of different loss components during training, allowing for adaptive optimization of our multi-objective function. 

\textbf{Neighbourhood Sampling and Subgraph Generation}
In large-scale heterogeneous graphs, processing the entire graph for each recommendation task can be computationally expensive. 
We focus on the most relevant parts of the graph and utilize a carefully designed edge splitting strategy for training, validation, and testing, balancing between information propagation and model supervision. During training, negative edges are generated on-the-fly which helps in efficient learning of edge distinction.

The multi-hop sampling strategy (\autoref{fig:framework} block d), combined with dynamic negative sampling during training and incorporating random walk information, allows the model to explore a broader range of graph structures while focusing on the most informative negative examples. This also enables the model to effectively learn from sparse interaction graphs, capturing complex relationship between monitors, dimensions, and metrics, while maintaining computational feasibility. 

\textbf{Computational complexity}: The sparse graph attention mechanism in the proposed method has linear time complexity with respect to the number of edges ($\mathcal{O}(\lvert E \rvert)$). The attention mechanism for the RWA module has a complexity of $\mathcal{O}(\lvert N \rvert ML^2) = \mathcal{O}(\lvert N \rvert)$, where $\lvert N \rvert$ is the number of nodes, $M$ is the number of paths and $L$ is the path length. Thus the total complexity is $\mathcal{O}(\lvert E \rvert)$ which can be reduced to $\mathcal{O}(\lvert N \rvert)$ (for our case of bounded degree graph). 
This is further reduced to sub-linear complexity with the sub-graph sampling methods employed, thus making the method scalable to production scale graphs.

\section{Experiments}
To evaluate the effectiveness of our proposed framework, we conduct a series of experiments on the task of dimension recommendation for monitors in cloud environments. This section details our experimental setup, evaluation metrics, baseline comparisons, and results analysis. Specifically we aim to answer the following research questions (RQs):
\begin{itemize}[leftmargin=*]
    \item \textbf{RQ1}: Is our method able to effectively predict relevant dimensions for a given monitor?
    \item \textbf{RQ2}: Does the optimization framework help in effectively ranking the relevant dimensions?
    \item \textbf{RQ3}: Does the addition of paths using random walks help alleviate the dimension sparsity problem?
    \item \textbf{RQ4}: Does our method scale with the data?
    \item \textbf{RQ5}: Does our method generalize to other datasets?
\end{itemize}
\subsection{Experiment Setup}
\subsubsection{Datasets}

Our experiments utilize a heterogeneous graph dataset representing a complex cloud monitoring system. The dataset comprises of three types of entities (nodes) and three types of relationships (edges). It captures the intricate interactions between monitors, dimensions, and metrics in a cloud environment. 
The main statistics of the dataset are summarized in
\autoref{tab:dataset}.
This graph structure effectively represents the complex relationships in cloud monitoring systems, where monitors are associated with specific dimensions and emit various metrics, while metrics themselves are characterized by multiple dimensions.

\begin{table}[t]
       \caption{Cloud monitoring Dataset}
    \label{tab:dataset}
    \centering
    \resizebox{\columnwidth}{!}{
    \begin{tabular}{cc}
     \textbf{Node} & \textbf{Edge} \\
 \hline \\
        \makecell{\# monitors: 18291 \\ \# metrics: 4623 \\ \# dimensions: 8356} & \makecell{\# monitor, associated\_with, dimension: 52148 \\ \# metric, has, dimension: 109213 \\ \# monitor, emits, metric: 52148}\\
    \end{tabular}
    }
\end{table}

\textbf{Feature Representation:} To capture the semantic information of monitors and dimensions, we employ feature embeddings generated using a state-of-the-art language model: 
1)  Monitor Features: Represented by the embeddings of the metric names associated with each monitor. 
and 2) Dimension Features: Represented by the embedding of the dimensions names. 

Both feature embeddings are generated using the ``E5'' embedding model, a general-purpose model trained through contrastive learning \citep{wang2022text}. This approach allows us to capture rich semantic information from the textual descriptions of metrics and dimensions, enabling our model to understand and utilize the contextual relationships between different entities in the cloud monitoring system. 
\subsubsection{Training Details} We train DiRecGNN using \(L=3\) message passing layers with hidden channels size of \(256\) and output channels size of \(128\). We use the Adam optimizer with a learning rate of \(0.001\) and weight decay of \(10^{-5}\). To adapt the learning rate during training, we employed a learning rate scheduler which reduces the learning rate by half if the validation loss doesn't improve for \(5\) consecutive epochs. We run our training for a maximum of \(100\) epochs, with early stopping implemented to prevent over-fitting. We use a patience of \(10\) epochs for early stopping. The edge set is divided into training (\(80\%\)), validation (\(10\%\)), and test (\(10\%\)) sets. Within the training set, we use \(70\%\) of edges for message passing, and \(30\%\) for supervision. For evaluation, we generate fixed negative edges with a ratio of \(2:1\) (negative to positive). During training, we generate negative edges dynamically using on-the-fly negative sampling. We sample multiple hops from both ends of a link to create subgraphs. We use a negative sampling ratio of \(2.0\) and employ a batch size of \(128\) during training. 
For the RWA module, we sample paths from monitor to dimension with lengths of $\{2,6,10\}$.

\noindent \textbf{Baselines and Metrics} :We evaluate the performance of our model using the evaluation metrics of Hit Ratio (HR@k), Mean Reciprocal Rank (MRR), Normalized Discounted Cumulative Gain (NDCG@k), Recall@k. We compare against traditional collaborative filtering \cite{collaborative_filtering_Wang_2021}, non-graph based MLP \cite{MLP_Rumelhart1986} (using only the text features), standard graph methods such as SAGEConv with mean/max pooling (SAGE(v1), SAGE(v2)) \cite{hamilton2017inductive}, GATConv (GAT) \cite{velickovic2017graph}, Transformer (T) \cite{shi2020masked} and the methods for heterogeneous graphs such as HGT \cite{HGT}, HAN \cite{HAN}, HetGNN \cite{HetGNN}.

Our proposed model was evaluated in three configurations:
\begin{itemize}[leftmargin=*]
    \item \textbf{Attention-alignment Loss} (T + AL): Our model featuring a custom TransformerConv with multi-head attention and integrated layer-wise attention-alignment loss.
    \item \textbf{Attention-alignment Loss + Ranking Loss} (T + AL + RL): Our model featuring a custom TransformerConv with integrated layer-wise attention-alignment loss and ranking optimization.
    \item \textbf{DiRecGNN: Attention-alignment Loss + Ranking Loss + RWA} (T + AL + RL + RWA): Our complete model featuring a custom TransformerConv with integrated layer-wise attention-alignment loss, ranking optimization along with the attention over random walk paths (RWA).
\end{itemize}
\subsection{Results (RQ1)}

\begin{table}[h!]
  \caption{Proposed framework outperforms baselines demonstrating relative gains of 55.3\%, 69.2\%, and 43.02\% in Hit-Rate@1, NDCG@k, and Recall@5 respectively with respect to the best baseline.} 
    \label{tab:my_label}
    \centering
    \resizebox{\columnwidth}{!}{
    \begin{tabular}{c c c c c c c c c}
      \textbf{Metric} & \textbf{HR@1} & \textbf{HR@3} & \textbf{HR@5} & \textbf{MRR} & \textbf{N@k} & \textbf{R@1} & \textbf{R@3} & \textbf{R@5} \\
        \hline \hline
        CF & 0.230 & 0.149 & 0.116 & 0.391 & 0.217 & 0.117 & 0.310 & 0.438\\
        MLP & 0.312 & 0.186 & 0.128 & 0.464 & 0.307 & 0.184 & 0.392 & 0.492\\
        \hline
        SAGE(v1) & 0.383 & 0.186 & 0.127 & 0.499 & 0.328 & 0.218 & 0.379 & 0.474\\
        SAGE(v2) & 0.291 & 0.154 & 0.111 & 0.414 & 0.262 & 0.165 & 0.323 & 0.398\\
        GAT        & 0.355 & 0.185 & 0.18 & 0.487 & 0.323 & 0.213 & 0.397 & 0.493 \\
        HGT        & 0.396 & 0.185 & 0.131 & 0.510 & 0.356 & 0.228 & 0.402 & 0.507 \\
        HAN        & 0.348 & 0.194 & 0.131 & 0.485 & 0.315 & 0.202 & 0.407 & 0.500 \\
        HetGNN        & 0.375 & 0.173 & 0.121 & 0.492 & 0.321 & 0.224 & 0.400 & 0.502 \\
        T & 0.331 & 0.188 & 0.134 & 0.481 & 0.306 & 0.178 & 0.399 & 0.523 \\
        \hline
         \makecell{T + AL} & 0.547 & 0.238 & 0.157 & 0.650 & 0.500 & 0.21 & 0.561 & 0.655 \\
         \makecell{+ RL} & 0.573 & 0.246 & 0.159 & 0.672 & 0.525 & 0.342 & 0.592 & 0.675 \\
         \makecell{+ RWA} & \textbf{0.597} & \textbf{0.265} & \textbf{0.173} & \textbf{0.714} & \textbf{0.555} & \textbf{0.355} & \textbf{0.649} & \textbf{0.748} \\
    \end{tabular}
    }
\end{table}

We observe from the results that the MLP model outperforms the user based collaborative filtering indicating the usefulness of the text features. Further we note the best GNN baseline (SAGE-conv) performs better than the vanilla MLP over the features, demonstrating that incorporating the graph structure into the learning helps in addition to the node features.
All variants of our proposed model significantly outperform the baselines across all metrics.  
The full model with both attention-alignment and ranking losses along with the Random Walk Attention (RWA) module shows the best overall performance. 
Our full model achieves HR@1 of 0.597, which is a 55.8\% improvement over the best baseline (SAGEConv + Mean at 0.383),  indicates a substantial enhancement in the ability to recommend the most relevant dimension as the top choice. The NDCG@k score of 0.555 for our full model, compared to 0.328 for the best baseline, represents a 69.2\% improvement. The results suggests that our model not only recommends relevant dimensions but also ranks them more effectively, answering RQ1. Our model shows significant improvements in Recall@k, particularly at higher k values. The Recall@5 of 0.748 for our full model compared to 0.523 for
TransformerConv indicates a 43.02\% improvement in retrieving relevant dimensions within the top 5 recommendations. 
The introduction of attention-alignment loss alone leads to substantial improvements across all metrics compared to the baselines. This underscores the importance of encouraging sparse attention patterns in the model for the sparse cloud monitoring entity graph.
The addition of ranking loss to attention-alignment loss results in further improvements, particularly in HR@1 and MRR. This highlights the effectiveness of our multi-faceted loss function in optimizing both accuracy and ranking quality.

DiRecGNN  uses a multi-head attention mechanism, which allows it to capture the relative importance of different neighbor types more effectively than baselines. On the other hand, while ``TransformerConv'' (T) uses attention, it may struggle to distinguish between different types of relationships as effectively as DiRecGNN.
The attention-alignment loss in the framework encourages it to capture relevant sparse aspects of the data, whereas the baselines do not have an explicit mechanism to encourage such sparsity in their representations.
The ranking loss from DiRecGNN directly optimizes for ranking quality. In contrast, the baselines are typically trained with binary classification objectives, which may not directly optimize for ranking quality.
The Random Walk Attention (RWA) module enables the model to capture long range global dependencies. In most of the baselines this information may be lost due to the homophilic nature of the aggregation.

\subsection{Ablation Study}
To understand the individual contributions of our model's key components, we conduct ablation studies focusing on the RWA module and on attention-alignment loss and ranking loss. 

\begin{figure}[h!]  
\centering
\begin{subfigure}[b]{0.22\textwidth}
       \centering
\includegraphics[width=\linewidth]{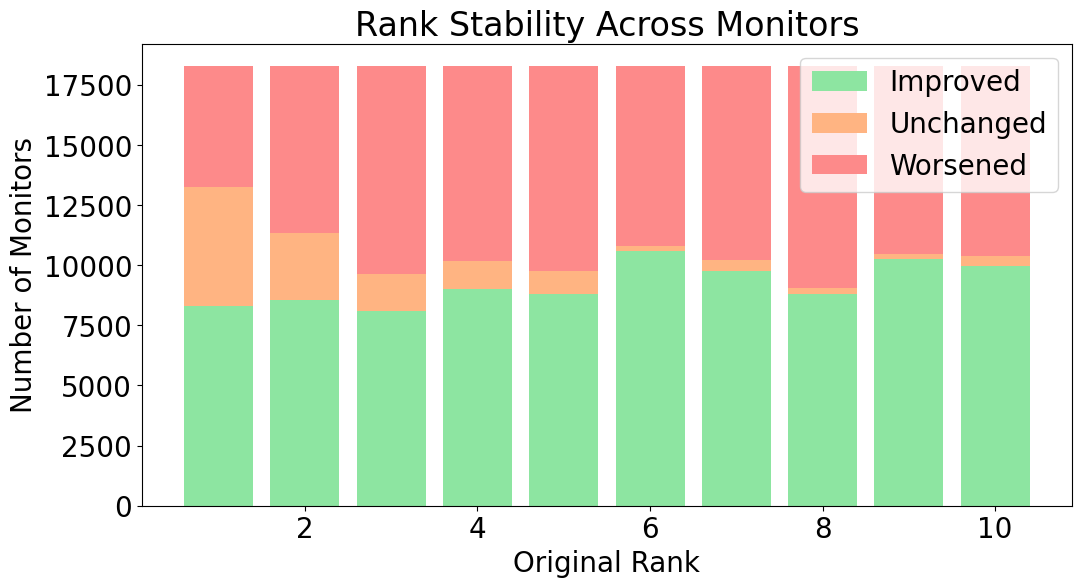}%
\caption{ }
\label{subfig: rankstability1}
\end{subfigure}
\begin{subfigure}[b]{0.22\textwidth}
       \centering
\includegraphics[width=\linewidth]{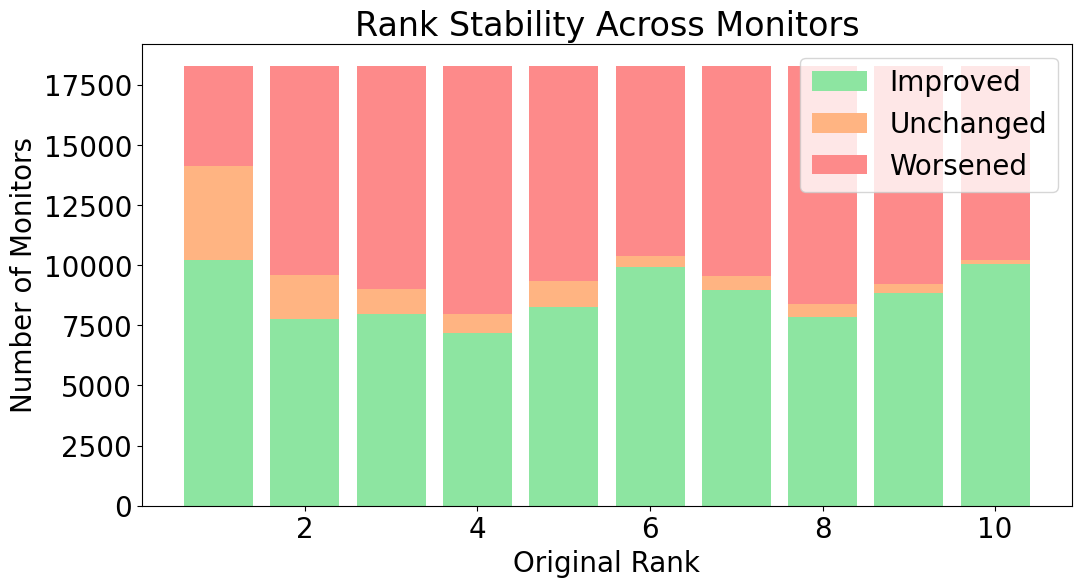}%
\caption{ }
\label{subfig: rankstability2}
\end{subfigure}
    \caption{a) Impact of Ranking Loss on model performance, and b) Impact of Attention-alignment Loss on model performance. Losses improve overall ranking.}
    \label{fig: rankstability}
\end{figure}

\noindent \textbf{Ablation on losses (RQ2):}
In this study (cf. \autoref{fig: rankstability}), we compare three model variants: 1) Base model (without attention-alignment, ranking loss and RW), 2) Model with only ranking loss and 3) Model with only attention-alignment loss.
We aggregate the changes in the ranks of the recommended dimensions across different monitors and the performance of these variants using rank stability plots, which visualize change in the relevance of top ranked dimensions across different model configurations. 
Overall quality of recommended dimensions improved with the
addition of the ranking loss and attention-alignment loss. 

\begin{figure}[h]
    \centering
     \begin{subfigure}[b]{0.15\textwidth}
     \centering
\includegraphics[width=\linewidth]{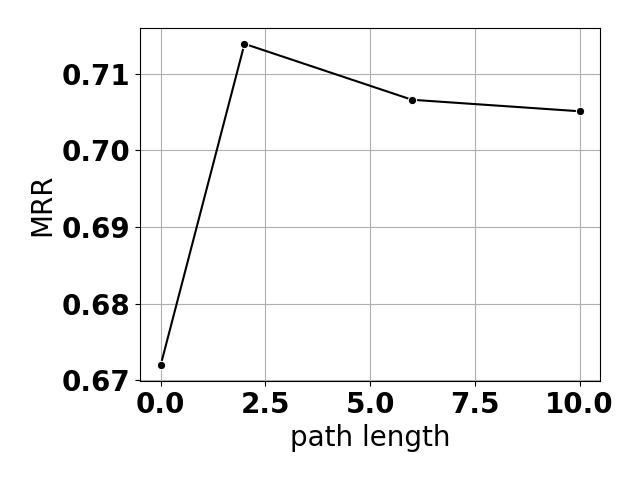}
        \caption{}
           \label{subfig:pl_vs_results}
    \end{subfigure}%
    \begin{subfigure}[b]{0.15\textwidth}
         \centering
\includegraphics[width=\linewidth]{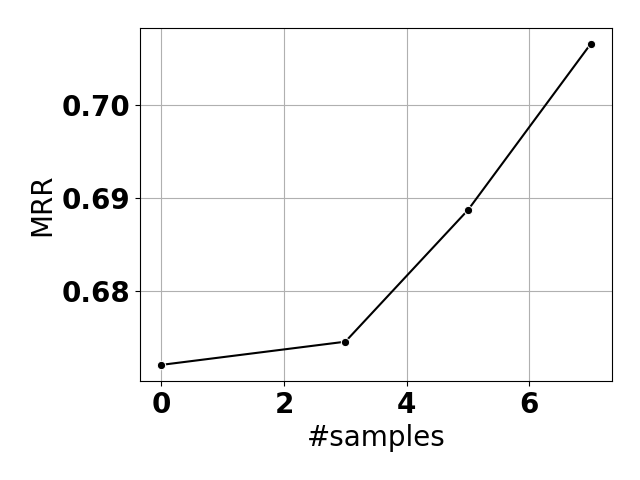}
    \caption{}
           \label{subfig:ns_vs_results}
    \end{subfigure}
    \begin{subfigure}[b]{0.15\textwidth}
         \centering
\includegraphics[width=\linewidth]{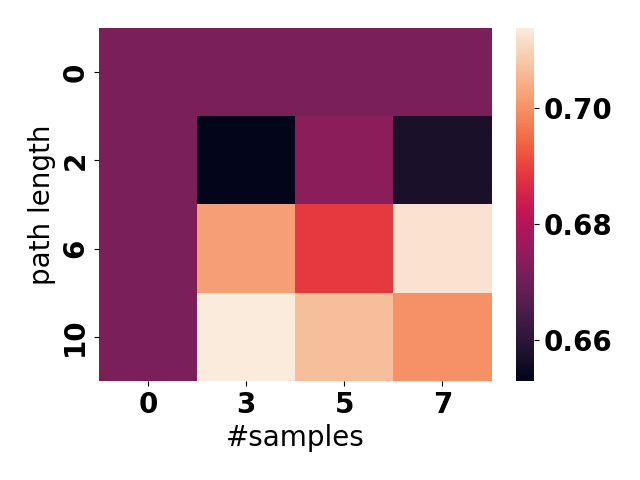}
    \caption{}
           \label{subfig:ns_vs_pl_results}
    \end{subfigure}
\caption{Ablation study on the RWA module: a) The variation of results (MRR) w.r.t. path length b) The variation of results (MRR) w.r.t number of samples c) The heatmap showing variation in results (MRR) with change in path length and $\#$ samples ($\uparrow$ trend with $\#$samples \& path length).}
\label{fig:rwa_ablation}
\end{figure}

\begin{figure}[ht]
    \centering
     \begin{subfigure}[b]{0.24\textwidth}
     \centering
\includegraphics[width=\linewidth]{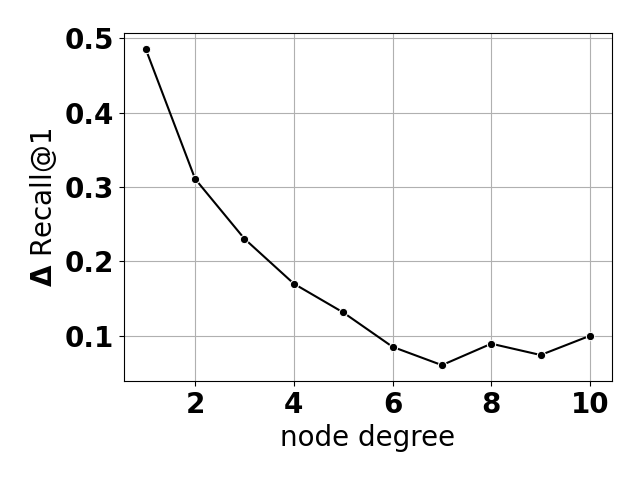}
        \caption{}
           \label{subfig:sparsity_vs_perf}
    \end{subfigure}%
    \begin{subfigure}[b]{0.24\textwidth}
         \centering
\includegraphics[width=\linewidth]{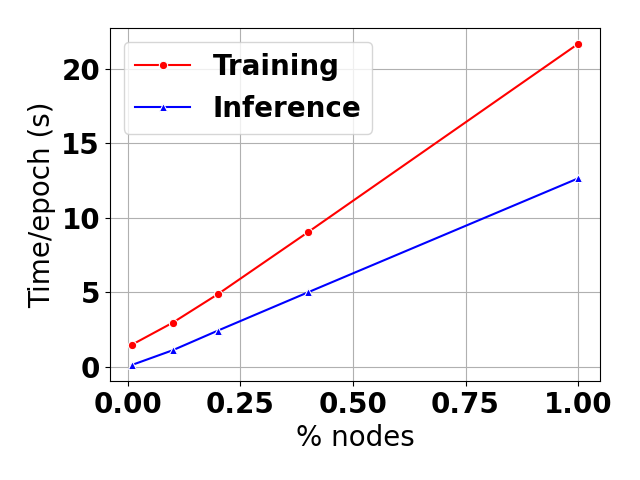}
    \caption{}
           \label{subfig:scalability}
    \end{subfigure}
\caption{a) The performance gains ($\uparrow$ with sparsity) by using RWA module w.r.t node degree b) Plot of percent nodes sampled vs time taken indicating linear scaling.}
\label{fig:sparsity_scalability}
\end{figure}

\textbf{Ablation on RWA module:}
Here we focus on the hyperparameters of the RWA module.
Figure \ref{fig:rwa_ablation} shows the results with variation in path length (No. samples=5), variation in number of samples (path length=10) and the landscape of the results with varying path length and number of samples. We can observe that as the path length increases the performance shows an improving trend indicating that the meta paths sampled using random walks indeed helps in capturing the long range dependencies benefiting the task. Similarly, we see improvements with more samples indicating more samples may help with the method being robust to noise. There is an exception for the case where the path length is 2 as this homophilic (neighborhood) information is already effectively captured by the message passing framework.

\textbf{Study on performance gains with sparsity (RQ3):} We study how the performance of the model varies with the degree of the node. Figure \ref{subfig:sparsity_vs_perf} shows the performance gain w.r.t. monitor node degree. We observe the difference in results are more pronounced in the lower node degrees implying that incorporating long range paths in the model enhances performance when the data is sparse.

\textbf{Scalability study (RQ4):} We also study how the model scales with number of nodes. The time taken for training and inference scales linearly with the number of nodes (cf. \autoref{subfig:scalability}). This shows that our method can scale to very large graphs in the real world.

\begin{table}[h!]
\caption{Node classification on benchmark dataset.}\label{tab:NC}
\small
\begin{tabular}{ccc}
\toprule
                    & \multicolumn{2}{c}{DBLP}                \\ \midrule
                    & Macro-F1              & Micro-F1        \\ \midrule
RGCN	&	91.52$\pm$0.50	&	92.07$\pm$0.50\\
HAN	&	91.67$\pm$0.49	&	92.05$\pm$0.62\\
GTN	&	93.52$\pm$0.55	&	93.97$\pm$0.54\\
RSHN	&	93.34$\pm$0.58	&	93.81$\pm$0.55\\
HetGNN	&	91.76$\pm$0.43	&	92.33$\pm$0.41\\
MAGNN	&	93.28$\pm$0.51	&	93.76$\pm$0.45\\
HetSANN	&	78.55$\pm$2.42	&	80.56$\pm$1.50\\
HGT	&	93.01$\pm$0.23	&	93.49$\pm$0.25\\ 
GCN	&	90.84$\pm$0.32	&	91.47$\pm$0.34\\
GAT	&	93.83$\pm$0.27	&	93.39$\pm$0.30\\ 
Simple-HGN & \underline{94.01$\pm$0.24} & \underline{94.46$\pm$0.22} \\ \midrule
Ours & \textbf{95.03$\pm$0.42} & \textbf{95.29$\pm$0.35} \\
\bottomrule
\end{tabular}
\end{table}

\textbf{Generalizability to Open Source Datasets (RQ5)}
We study how our method would generalize to open source datasets and across tasks of node classification and link prediction. For this we select two datasets one for each task namely \footnote{http://web.cs.ucla.edu/~yzsun/data/}{DBLP} for node classification and LastFM \cite{lastfm} for link prediction. To ensure fair comparison, we adopt the same settings as in \cite{lv2021reallymakingprogressrevisiting}. The results are presented in table \ref{tab:NC}, \ref{tab:recom_benchmark}. We see that our method performs comparably to the baseline methods affirming the generalizability of our method.

\begin{table}[h!]
\caption{Knowledge-aware recommendation benchmark. We adopt the baselines from \cite{lv2021reallymakingprogressrevisiting}.}
\label{tab:recom_benchmark}
\centering
\footnotesize
\begin{tabular}{ccc}
\toprule
       & \multicolumn{2}{c}{LastFM} \\ \midrule
        & recall@20       & ndcg@20\\ \midrule
KGCN	&	0.0819$\pm$0.0002	&	0.0705$\pm$0.0002\\
KGNN-LS	&	0.0806$\pm$0.0003	&	0.0695$\pm$0.0002\\
KGAT	&	0.0877$\pm$0.0003	&	0.0749$\pm$0.0003\\
KGAT$-$	&	0.0890$\pm$0.0002	&	0.0762$\pm$0.0002\\
Simple-HGN	&	\underline{0.0917$\pm$0.0006}	&	\underline{0.0797$\pm$0.0003} \\ \midrule
Ours	&	\textbf{0.1026$\pm$0.0002}	&	\textbf{0.2051$\pm$0.0002}\\ 

\bottomrule
\end{tabular}
\end{table}

\begin{figure}[t]  
\centering
\begin{subfigure}[b]{0.22\textwidth}
       \centering
\includegraphics[width=\linewidth]{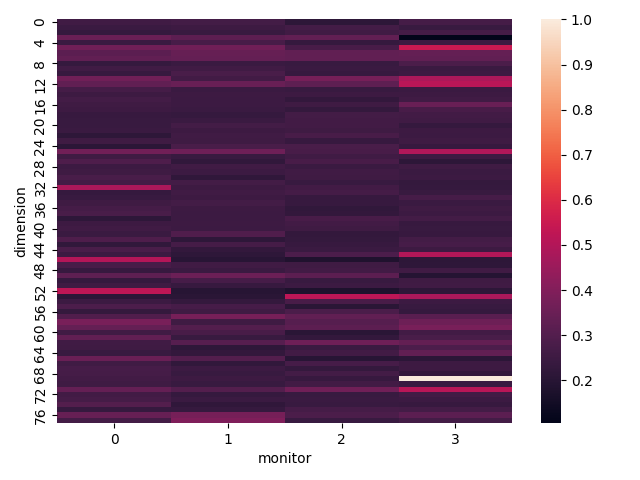}%
\caption{ }
\label{subfig: dim_rev_monitor_nodiv_bucket4}
\end{subfigure}
\begin{subfigure}[b]{0.22\textwidth}
       \centering
\includegraphics[width=\linewidth]{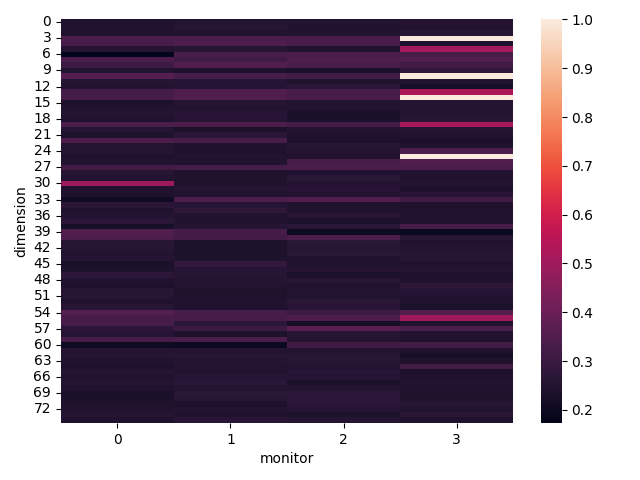}%
\caption{ }
\label{subfig: dim_rev_monitor_div_bucket4}
\end{subfigure}

\begin{subfigure}[b]{0.22\textwidth}
       \centering
\includegraphics[width=\linewidth]{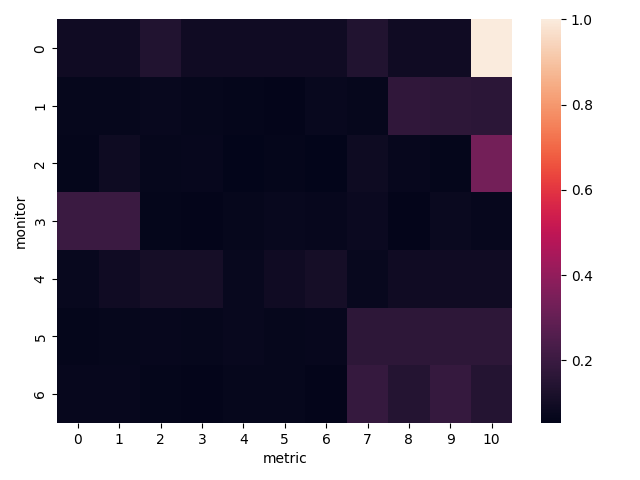}%
\caption{ }
\label{subfig: monitor_emits_metric_nodiv_bucket11}
\end{subfigure}
\begin{subfigure}[b]{0.22\textwidth}
       \centering
\includegraphics[width=\linewidth]{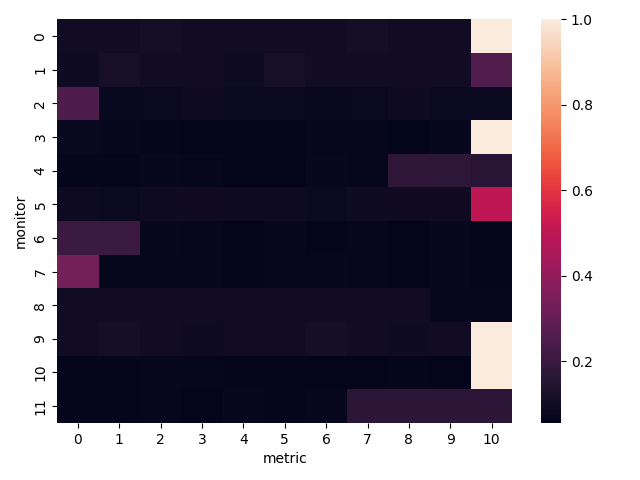}%
\caption{ }
\label{subfig: monitor_emits_metric_div_bucket11}
\end{subfigure}

\begin{subfigure}[b]{0.22\textwidth}
       \centering
\includegraphics[width=\linewidth]{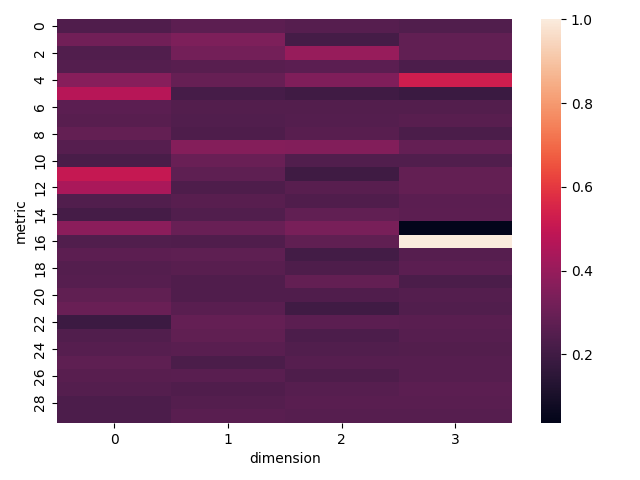}%
\caption{ }
\label{subfig: metric_has_dimension_nodiv_bucket4}
\end{subfigure}
\begin{subfigure}[b]{0.22\textwidth}
       \centering
\includegraphics[width=\linewidth]{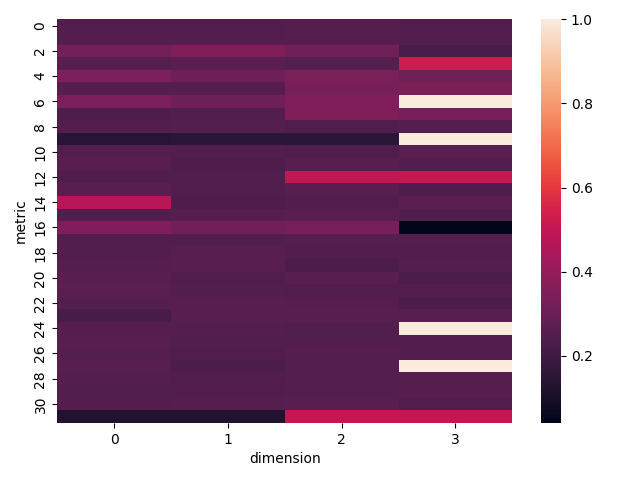}%
\caption{ }
\label{subfig: metric_has_dimension_div_bucket4}
\end{subfigure}
    \caption{a) Attention map between monitor and dimension nodes without alignment loss b) Attention map between monitor and dimension nodes with alignment loss c) Attention map between monitor and metric nodes without alignment loss d) Attention map between monitor and metric nodes with alignment loss e) Attention map between metric and dimension nodes without alignment loss f) Attention map between metric and dimension nodes with alignment loss }
    \label{fig:attention_heatmap}
\end{figure}

\textbf{Visualization of learnt attention maps with the Attention-alignment loss}
We further study the attention maps learnt using the alignment loss and without the loss. With the alignment loss we expect the attention weights to be concentrated across heads respecting the sparse connectivity pattern. Figure \ref{fig:attention_heatmap} shows the attention heatmap between sampled nodes of each edge type with the attention weights learnt without the aligment loss on left and with the loss on right. The attention weights are averages across the heads. We see the attention is more distributed without the loss whereas with the loss it is more concentrated on particular entities. The behaviour could be due to different heads learning distinct weights for different entities without the loss, which causes a diffused attention pattern on average. This validates our claim that the alignment loss respects the sparse data connectivity between entities in the graph.
\section{Lessons Learnt from Production}

In this section, we summarize the key findings learned during the deployment of the monitor framework in production. We conducted a study with a subset of approximately 30 users from product teams (service owners) who use our recommendations. The study aimed to identify the major challenges encountered during monitor creation and updating, evaluate the effectiveness of the recommendations, and determine areas for improvement.

\noindent It was observed that most services update monitors on a weekly basis. The main challenges were determining the subset of dimensions to monitor and setting appropriate thresholds for raising alerts. We addressed the former in this work and plan to target the latter in future efforts. We also learned that recommendations were more actionable when accompanied by explanations about which similar service is monitoring the recommended dimensions. Additionally, one user requested end-to-end automation of monitor creation and the display of recommendations using prompt boxes. Overall, users perceived the feature as helpful during both the creation and updating of monitors, giving it an average relevance rating of 4.5 out of 5.

\begin{figure}[]
    \centering
    \includegraphics[width=0.95\linewidth]{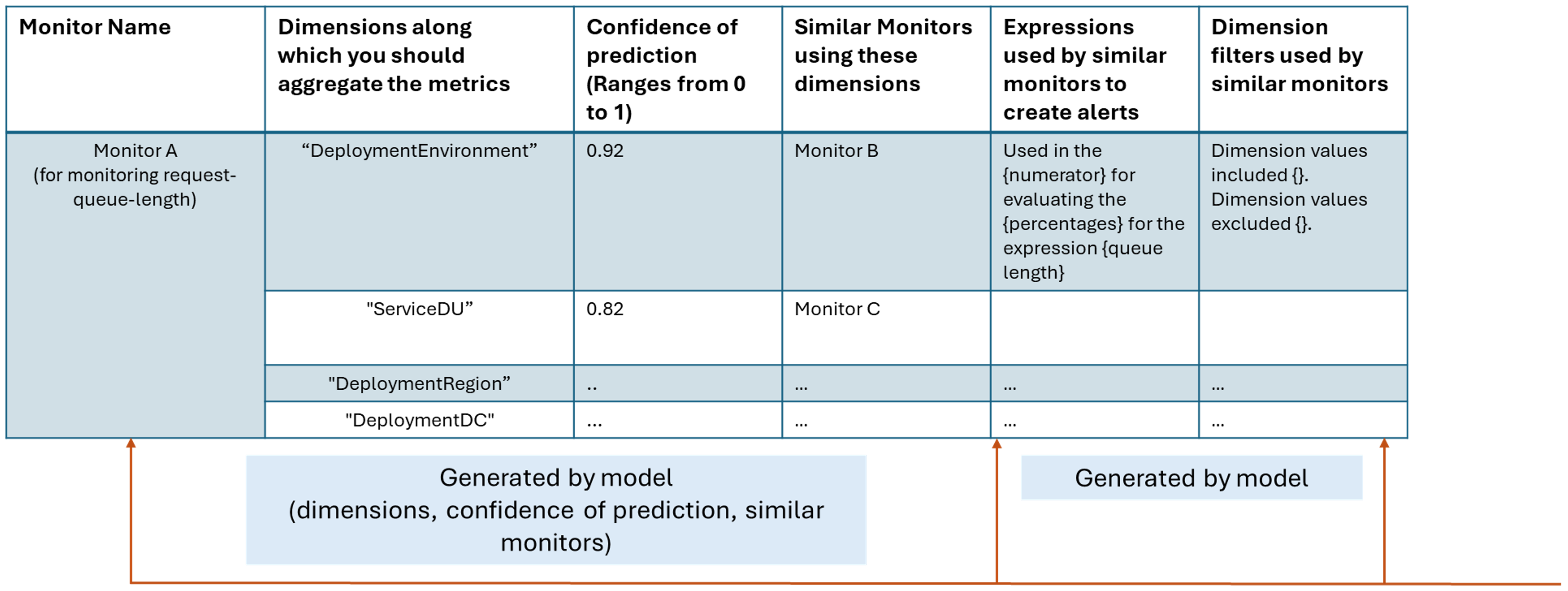}
    \caption{Example recommendation of dimensions for a monitor as it would appear in the system UI. We provide the dimension recommendations along with the scores to help service owners better decide. Moreover based on feedback from user study we provide explanations into similar monitors that use the recommended dimensions.}
    \label{fig:system_ui}
\end{figure}

\subsection{Setup for User Study}
To select study participants, we generated a list of \company engineers who modified monitors between June and December 2024. We filtered out those who modified monitors fewer than 10 times, then randomly sampled 30 engineers for interviews. Of these, 10 agreed to participate, resulting in a response rate of 33\%. Over six months, the participants made a minimum of 74 and a maximum of 797 modifications. We conducted 10-minute structured interviews with these 10 participants.

During the interviews, we asked participants how often they modified their monitors and recorded any issues encountered during the process. We also inquired about the challenges they faced when creating or updating monitors. Next, we introduced the recommendation feature, asked whether it was helpful, and requested a usefulness rating. We asked, ``At which stages of the monitor lifecycle do you believe the dimensions recommendation would be useful?'' Furthermore, we asked, ``What kind of user experience would you prefer when these recommendations are presented via the monitor creation portal, and what other overheads does your team face while using this solution?''
Below are the questions and results of the study: \\
\noindent \textbf{Q1. How often do you create/update monitors?
} Among the 10 participants, 5 reported that they modify monitors on a weekly basis, 4 reported doing so on a monthly basis, and 1 reported that they modify monitors on a yearly basis.

\noindent \textbf{Q2.Do you think a structured, explainable recommendations for monitor configurations, such as dimensions, would be helpful during monitor creation/modification?}
All participants agreed that this feature is useful. For example, P1 stated, \textit{"Yes, based on the selected metric and the monitor's purpose or expression, it is good to predict which dimensions one should use."} P5 commented, \textit{"Such recommendations will be helpful. We have done it manually in the past."} P10 added, \textit{"This will save a lot of time."}

\noindent \textbf{Q3: At what stages of the monitor life cycle do you feel the dimensions recommendation is useful?
}
All participants indicated that this feature is useful during monitor creation, and 4 out of 10 also noted that it is beneficial for resolving buggy monitors.

\noindent \textbf{Q4: On a scale of 1 to 5, how useful are the recommendations
} 
 We observed an average rating of 4.5 out of 5 for the usefulness/relevance of the recommendations provided. Three participants reported that the monitor recommendations were extremely useful, while seven stated that they were somewhat useful. One participant responded neutrally, and no participant indicated that the recommendations are not useful.

 \noindent \textbf{
Q5: What kind of additional user experience would you prefer while we surface these recommendations via monitor creation portal? }

\textit{P1:``I expect end-to-end automation''}

\textit{P2: ``A prompt box with clear and concise directions for each stage''}

\textbf{Q6: 
What are some additional overheads faced when integrating this solution into your service via monitor creation portal?}
While nine out of ten participants were satisfied with the current options and did not face any additional overhead, one participant provided the following input:

\textit{P9: ``Users needed time to adapt to the new recommendation feature, which required additional training and support documentation.''}

\myparagraph{Example of System Output}\label{sec:system_ui}
In this section we give an example of how the recommendation from the system would look like to the user. In the figure \ref{fig:system_ui} we see an example recommendation for a monitor (Monitor A) which is built for monitoring the request queue length. The recommended dimensions are provided along with the recommendation scores which help the user decide how confident the model is about the prediction. Moreover in order to provide some explainability to the recommendations, based on the user study, we also provide similar monitors that use these dimensions. In addition to the similar monitor names we also provide the mathematical expressions in which these dimensions are used along with any filters to dimension values (that inform whether to exclude or include certain dimension values from the alerting logic).

\section{Conclusion}
We introduce the problem of dimension recommendation for cloud monitoring systems. To solve it, we formulate a heterogeneous monitor entity graph and propose a novel recommendation framework, DiRecGNN, which leverages the available set of node attributes and multi-type interactions to overcome the challenges of structural and interaction sparsity and long range dependencies in the network.
DiRecGNN incorporates an attention-alignment loss, edge-aware message passing, and multi-head attention mechanisms over random walk paths inspired by transformer architectures. Experiments on the monitor-entity dataset demonstrate significant improvements in hit-rate, mean reciprocal rank, and recall over baseline approaches. The proposed framework presents a promising approach for addressing the recommendation problem in cloud settings with sparse interaction data.

\bibliographystyle{ACM-Reference-Format}
\bibliography{base}

\end{document}